\documentclass[preprint,12pt]{elsarticle}
\usepackage[final]{microtype}   % better spacing + subtle font expansion
\usepackage{subcaption}
\usepackage{booktabs}
\usepackage{placeins}
\usepackage{threeparttable}
\journal{Journal of Biomedical Informatics}
\usepackage{amssymb}
\usepackage{ulem}
\usepackage{xcolor}
\usepackage{float}
\begin{document}
\begin{frontmatter}
%% Title, authors and addresses
\title{Latent Knowledge as a Predictor of Fact Acquisition in Fine-Tuned Large Language Models}
\author{Daniel B. Hier} %% Author name
%% Author affiliation
\affiliation{organization={Department of Neurology and Rehabilitation, University of Illinois at Chicago},%Department and Organization
            addressline={912 S. Wood Street}, 
            city={Chicago},
            postcode={60612}, 
            state={IL},
            country={USA}}
\author{Tayo Obafemi-Ajayi} %% Author name
%% Author affiliation
\affiliation{organization={Engineering Program, Missouri State University},%Department and Organization
            addressline={901 S. National Avenue}, 
            city={Springfield},
            postcode={65897}, 
            state={MO},
            country={USA}}

\begin{abstract}
\textbf{Objective:} Large language models encode biomedical facts with uneven strength after pretraining. While some facts are encoded in model parameters but inaccessible under deterministic decoding (\textit{latent knowledge}), others are minimally represented. We fine-tuned a large language model to learn facts about term–identifier pairs from the Human Phenotype Ontology (HPO) and the Gene Ontology (GO). We hypothesized that latent knowledge would predict the rate at which facts are acquired during fine-tuning and which facts would be acquired by generalization.

\textbf{Methods:} Llama-3.1-8B-Instruct was fine-tuned on 800 HPO term–identifier pairs and 400 GO term–identifier pairs. An additional 400 GO term-identifier pairs were withheld to assess generalization. Fact acquisition was modeled as a time-to-event process. Latent knowledge encoded in model parameters about term--identifier pairs was identified by stochastic-decoding. Cox proportional hazards models were used to quantify predictors of the rate of acquisition of HPO facts, the generalization of facts about \textit{unseen} GO terms, and the degradation of previously correct GO term–identifier mappings.

\textbf{Results:} At baseline, Llama-3.1-8B-Instruct correctly recalled 2.8\% of the HPO term–identifier pairs. After 20 epochs of supervised fine-tuning, the model learned 71.9\% of the pairs. Latent knowledge was the strongest predictor of the rate of fact acquisition (hazard ratio 2.6, $p < 0.001$). Facts associated with latent knowledge were learned earlier, at higher peak rates, and converged sooner. Identifier frequency and annotation counts in the biomedical literature had smaller but significant effects; raw term frequency did not. For GO, fact generalization to \textit{unseen} term–identifier pairs was uncommon (5.8\%) but more likely for terms with latent knowledge. Degradation of previously correct GO mappings was more likely for \textit{unseen} terms (not included in the training regime) than for \textit{seen} terms (included in training), indicating a protective effect of training.

\textbf{Conclusion:} Latent knowledge acquired during pretraining influences the rate of fact acquisition during fine-tuning and predicts which \textit{unseen} ontology facts will generalize. By contrast, resistance to degradation depends on whether facts are reinforced during training. Modeling fact acquisition, generalization, and degradation as rate-based processes assists in identifying when fine-tuning is most efficient for injecting biomedical ontology facts into large language models.
\end{abstract}

%% Keywords

\end{frontmatter}
\section{Introduction}
\label{sec:introduction}
Although originally designed as text generators, large generative AI models are widely used as repositories of facts that can be queried on demand~\cite{wang2023survey, mousavi2025llms}. However, this use is limited by knowledge gaps~\cite{mousavi2025llms}. Because these models are trained on raw text rather than curated fact databases~\cite{ovadia_3bf77322, mecklenburg_cf5151d5}, the facts encoded in their parameters mirror the uneven distribution of facts in their training corpus. Since large language models are trained on tokens rather than on curated facts, frequently occurring  facts are encoded strongly, whereas rarely  encountered facts are encoded weakly and are harder to retrieve~\cite{kandpal2023large}. This \textit{long-tail} effect motivates ongoing work on scaling~\cite{lu2024scaling, kaplan2020scaling}, supervised fine-tuning~\cite{ghosal2024understanding}, retrieval-augmented generation~\cite{li2024role}, and knowledge editing~\cite{yi2025can}. Although fine-tuning can be used to teach both procedural skills and declarative facts~\cite{li2024meta}, our work focuses on fact acquisition, operationalized as learning the correct mapping between an ontology term and its unique identifier.

Prior work~\cite{hier2025prior, Hier2025failures} has established that the popularity of a fact in the pretraining corpus influences whether it becomes encoded in a model’s parameters. Facts associated with frequent textual mentions are more likely to be acquired, whereas rare facts often remain unknown to the model~\cite{kandpal2023large, ghosal2024understanding}.  For biomedical ontologies, the factual salience of a fact (how strongly it is represented in the model parameters) correlates with the frequency of its ontology terms and identifiers in biomedical literature such as PubMed Central, as well as with its frequency in curated annotation resources~\cite{hier2025prior, Hier2025failures}. However, stochastic-decoding studies have shown that some facts present in the model’s weights are inaccessible under deterministic decoding and constitute \textit{latent knowledge}~\cite{gekhman2025inside}. Facts differ both in whether they are represented at all and in how easily they can be retrieved.

There is growing evidence that the internal representations of facts formed during pretraining influence how models respond to fine-tuning. Facts that are salient in a model’s parameters are easier to strengthen, whereas low-salience facts require more supervision and are acquired more slowly~\cite{ovadia_3bf77322, mecklenburg_cf5151d5, wu2024finetunebench}. Long-tail facts—rare items with minimal support in the training corpus—are less likely to be encoded initially and remain difficult to learn, even with substantial fine-tuning~\cite{kandpal2023large}. Beyond these long-tail effects, recent work has identified \textit{ontology deserts}: groups of ontology terms that are structurally defined in the ontology but essentially absent from biomedical text and annotation corpora, and that therefore receive little or no representational capacity during pretraining~\cite{Hier2025failures}. Such desert regions pose a particular challenge for terminology-aware clinical language models, because they must be populated by remedial supervision after pretraining rather than by simply reinforcing partially encoded knowledge.
        
Most studies of factual acquisition during fine-tuning treat success as a binary outcome—either a fact is learned or it is not. Much less attention has been given to \textit{how quickly} different facts are acquired, which facts are lost during training, or which properties of a fact—such as latent knowledge or popularity in the corpus—govern its acquisition dynamics. The concept of \textit{generalization}, borrowed from skill-learning research where it refers to applying a learned skill to new inputs drawn from the same distribution, has no exact analog for fact acquisition~\cite{lu2024scaling,wu2025rote}. In particular, it is unclear how the training of models on one set of facts permits them to correctly answer questions about a distinct set of facts that were never encountered in training. In the setting of fact acquisition in general, and of learning ontology term–identifier mappings in particular, it is essential to distinguish among three outcomes:
\begin{itemize}
\item \textbf{Memorization}: improvement on \textit{seen} facts contained in the fine-tuning training set.
\item \textbf{Generalization}: improvement on \textit{unseen} facts withheld from the training set.
\item \textbf{Degradation}: loss of facts that were correct prior to fine-tuning.
\end{itemize}

Considerable effort has been devoted to understanding learning dynamics during pre-training and fine-tuning of LLMs. Most studies emphasize time to early stopping, time to convergence, or optimization of the number of training epochs~\cite{lu2024scaling, speicher2024memorisation, tirumala2022memorization, jiang2024unified, dodge2020fine, li2020train}. To our knowledge, none of these studies quantify facts gained or facts lost during fine-tuning as \textit{rates}. To address this gap, we treat fact learning as a time-to-event process and apply survival analysis methods to characterize fact acquisition during fine-tuning. In this framework, the hazard function derived from the survival function provides a natural notion of rate that captures the instantaneous velocity of fact acquisition during fine-tuning.

\textit{Latent knowledge} refers to factual information stored in the model’s parameters that is not reliably accessible through deterministic decoding but becomes detectable under stochastic sampling \cite{gekhman2025inside}. Intuitively, it reflects how much the model \textit{already knows} before fine-tuning begins. By modeling fact accumulation with Kaplan–Meier estimators and Cox proportional hazards models, we quantify how rapidly new facts are acquired, how often previously correct facts are degraded, and under what conditions models generalize to \textit{unseen} facts. We hypothesize that latent knowledge—information already represented in model weights—increases the rate of fact acquisition~\cite{gekhman2025inside,hier2025prior, pletenev2025much}, and further that the same signals predict which \textit{unseen} facts benefit from fine-tuning (generalize) and which previously correct facts resist degradation.

To test these hypotheses, we use ontology term–identifier pairs from two ontologies, the Human Phenotype Ontology (HPO) and the Gene Ontology (GO), as structured testbeds of biomedical facts. Each pair consists of a term and a unique identifier (\texttt{ataxia} $\rightarrow$ HP:0001251 in HPO or \texttt{mitochondrion} $\rightarrow$ GO:0005739 in GO). Although such mappings can be expressed as semantic triples (\texttt{term} $\rightarrow$ \texttt{has\_identifier} $\rightarrow$ \texttt{ID}), the essential fact is the term–identifier pair. These mappings are empirically verifiable, unambiguous, and widely used in biomedical text processing~\cite{krauthammer2004term, xu2024medical}. Even state-of-the-art models exhibit incomplete knowledge of HPO and GO pairs~\cite{Hier2025failures, do2025mapping, do2025balanced}. HPO serves as our primary testbed for modeling acquisition velocity, whereas GO allows systematic evaluation of fact generalization and degradation during fine-tuning. As fine-tuning progresses across epochs, it provides a natural timeline for time-to-event analysis: an event occurs when the model first outputs the correct identifier (acquisition) or first changes from correct to incorrect (degradation). In addition, for \textit{unseen} (held-out) GO terms, we measure the relatively uncommon cases in which a mapping transitions from incorrect to correct, yielding a fact-based example of generalization.
        
Beyond characterizing fine-tuning dynamics, this framework offers a general method for assessing how large language models acquire ontology-grounded biomedical knowledge. By separating memorization of \textit{seen} ontology facts, generalization to \textit{unseen} facts, and degradation of previously known facts, it clarifies when fine-tuning truly injects new knowledge versus reshapes or erodes existing representations.
    
The remainder of this paper is structured as follows. Section II describes our datasets and our methods for measuring the time required for the model to acquire, generalize, or degrade individual term–identifier pairs. Section III presents our results across HPO and GO, including acquisition velocities, degradation patterns, and predictors such as latent knowledge and popularity. Section IV discusses the implications of these findings for terminology-aware fine-tuning and outlines limitations and future directions.

\section{Methods}
\subsection{Terminology}
The term  \textit{deterministic decoding} implies argmax decoding, i.e. the process whereby the model selects the highest-probability next token. For all determinations of model accuracy, this study used deterministic decoding. \textit{Stochastic decoding} refers to repeated sampling from the model’s output distribution at a temperature of~1.0. Stochastic decoding was used to probe the base model for latent knowledge \cite{gekhman2025inside}.

The Llama-3.1-8B-Instruct model was considered to possess \textit{latent knowledge} about an ontology term (either HPO or GO) if at least one out of the 50 stochastic samples produced the correct identifier for that term. Terms that met this criterion were further stratified by the proportion of correct responses across the 50 samples (e.g., high versus moderate latent knowledge with a cutpoint at 10\% of stochastic samples correct).  \textit{Factual salience} denotes the strength with which a fact is represented in the model's parameters.  Operationally, the presence of latent knowledge serves as a marker of higher factual salience in the model \cite{ghosal2024understanding, gekhman2025inside}.

\subsection{Datasets}
We used term–identifier pairs from the \textit{Human Phenotype Ontology} (HPO) and the \textit{Gene Ontology} (GO) as the basic factual units to be learned during fine-tuning~\cite{robinson2015capturing, ashburner2000gene}. Each fact consists of a single ontology term and its unique identifier (e.g., \textit{ataxia} $\rightarrow$ HP:0001251 in HPO; \textit{nucleus} $\rightarrow$ GO:0005634 in GO). These mappings form unitary facts that are widely used across NLP and ontology-based applications in biomedicine.

The full HPO has 18,961 concepts. From this ontology, we sampled 800 term–identifier pairs, stratified to include both frequently occurring terms and long-tail terms that appear rarely in biomedical text. All 800 HPO facts were used as training examples (\textit{seen} terms) during fine-tuning.  GO has 39,354 terms across three hierarchies ( biological process, molecular function, and cellular component). We sampled 800 GO term–identifier pairs using an analogous frequency-stratified strategy. GO terms were divided into a training set (\textit{seen} terms) and a withheld evaluation set (\textit{unseen} terms), enabling a measurement of fact-based generalization. Terms that never appeared in the curated literature annotations—approximately 40\% of HPO and 54\% of GO terms~\cite{Hier2025failures}—were excluded to ensure that the sampled terms had at least a minimal grounding in human-generated text.

Term popularity \cite{kandpal2023large} was estimated using two sources that approximate the model’s pretraining exposure:
\begin{enumerate}
    \item  Identifier and term frequency in PubMed Central (PMC) full-text articles~\cite{hier2025prior, do2025balanced, do2025mapping}.
	\item Curated annotation frequency, using the number of disease annotations for each HPO term~\cite{kohler2017human, kohler2021human} or the number of gene/protein annotations for each GO term~\cite{hill2008gene}.
\end{enumerate} 
These measures served as proxies for popularity \cite{ni2025knowledge} in the pretraining data. The resulting stratified HPO and GO datasets provided controlled distributions of common, infrequent, and long-tail terms (exclusive of unused terms), enabling systematic evaluation of fact acquisition, generalization, and degradation across both ontologies.

\subsection{Fine-Tuning}
We fine-tuned the Llama-3.1-8B Instruct model~\cite{llama_together_ai} on ontology term–identifier pairs using a parameter-efficient approach. For HPO, all 800 sampled term–identifier pairs served as training examples (\textit{seen} terms). To increase robustness to prompt phrasing, we generated five paraphrased prompts for each fact, yielding 4,000 training instances per epoch.

For GO, the 800 sampled term–identifier pairs were split evenly into a training set (400 \textit{seen} terms) and a withheld evaluation set (400 \textit{unseen terms}). The withheld terms were not shown to the model during fine-tuning, enabling measurement of fact generalization to \textit{unseen} terms.

Fine-tuning used Low-Rank Adaptation (LoRA); rank $r=64$, scaling $\alpha=128$)~\cite{Lora_2021}, applied to all linear transformer layers. Optimization used a cosine learning-rate schedule (initial learning rate $1\times10^{-5}$, no warm-up, cycle length 0.5), batch size of 32, maximum gradient norm of 1, and zero weight decay. The \texttt{train-on-inputs} option was set to \texttt{auto}. Training was performed  and assessed for each epoch between 1 and 20 epochs.

For each ontology fact, we recorded the epoch at which the model first produced the correct identifier (\textit{acquisition}) or first changed from correct to incorrect (\textit{degradation}). These epoch-level transitions allowed us to treat fine-tuning as a time-to-event process and apply survival-analysis methods to quantify acquisition, degradation, and generalization.

\subsection{Analytic Framework and Visualization}
We modeled fine-tuning as a time-to-event process, where the event is the
first epoch in which the model produces the correct identifier for a given HPO or GO term. Beyond standard survival quantities (\(S(t)\), \(H(t)\), \(h(t)\))
\cite{kaplan1958nonparametric, collett2023modelling, kleinbaum2020survival}, we
characterized fact acquisition using rate-based measures derived from the
survival curve:
\begin{itemize}
    \item \(F(t) = 1 - S(t)\): the \textit{accumulation curve};
    \item \(V(t) = \frac{dF}{dt}\): the \textit{accumulation velocity};
\end{itemize}
Here, \(F(t)\) is the fact acquisition curve and \(V(t)\) captures the
instantaneous rate of new fact acquisition. Kaplan–Meier estimates of \(S(t)\) were used to derive \(F(t)\). Terms already known to the base model were right-censored at epoch~0. Confidence intervals used Greenwood’s formula with log–log transformation and were computed using the \texttt{lifelines} library (version 0.30.0).

Velocity was estimated using the first numerical derivative of a Gaussian-smoothed \(F(t)\) curve (\texttt{gaussian\_filter1d}, \texttt{scipy.ndimage}). Predictors of fact-acquisition rate were assessed using Cox proportional hazards models
\cite{cox1972regression}, yielding adjusted hazard ratios (HRs) with 95\%
confidence intervals. Continuous covariates were standardized; binary
covariates were left unscaled. Four covariates were included: latent knowledge and three popularity measures: (i) identifier frequency in PMC, (ii) term frequency in PMC, and (iii) curated disease-annotation count~\cite{hpo_annotations, hill2008gene}. All measures were
Laplace-smoothed and log-transformed. Analyses were implemented in Python (v3.10.14) using \texttt{lifelines}, \texttt{numpy}, \texttt{pandas}, and \texttt{matplotlib}. Cox model results were visualized using forest plots with HRs on a logarithmic scale and a reference line at HR~\(=1\). In all models, tests based on scaled Schoenfeld residuals indicated violations of the proportional hazards assumption, implying that hazard ratios were not constant over epochs. Accordingly, the reported HRs should be interpreted as time-averaged effects over the fine-tuning period rather than as epoch-specific effects \cite{in2019survival}.

\textit{Generalization} was assessed using  400 GO term–identifier pairs that had been from training (\textit{unseen}) Twenty-one pairs that were correct at baseline (epoch 0) were excluded from the generalization experiment. The remaining 379 \textit{unseen} terms were evaluated across fine-tuned models produced after each training epoch (epochs 1–20), where models were fine-tuned on the 400 \textit{seen} GO terms. For each \textit{unseen} term, if the model’s prediction changed from incorrect to correct at any given epoch, the event was recorded as a generalization occurring at that epoch. Kaplan–Meier estimators and Cox proportional hazards models were used to characterize generalization rates and their predictors, following standard survival-analysis procedures implemented in the \texttt{lifelines} Python library (version 0.30.0).

\textit{Degradation} was assessed using 22 GO terms from the training set (\textit{seen}) and 21 GO terms from the withheld set (\textit{unseen}) that had been correctly mapped to their GO identifiers at baseline (epoch 0). These initially correct terms were re-evaluated across each fine-tuned model produced after 1-20 epochs of training, where fine-tuning utilized only \textit{seen} GO terms for training. If model output for a term changed from correct to incorrect at any epoch, the event was recorded as a \textit{degradation} occurring at that epoch. Kaplan–Meier estimators and Cox proportional hazards models were used to characterize degradation risk and its predictors, following the above-described survival-analysis procedures.

\begin{figure}[htbp]
    \centering
    % Panel A: overall accumulation
    \begin{subfigure}{0.48\textwidth}
        \centering
        \includegraphics[width=\linewidth]{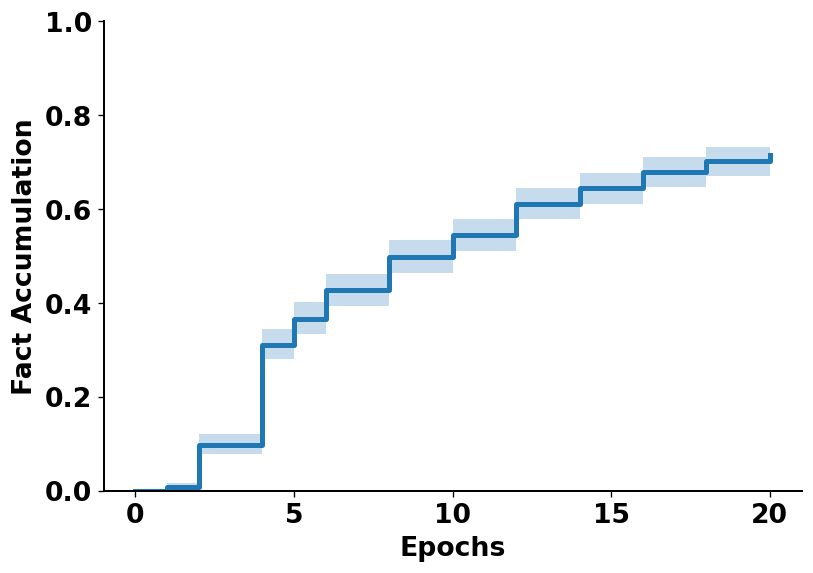}
        \caption{Overall HPO fact accumulation.}
        \label{fig:km-overall-hpo}
    \end{subfigure}
    \hfill
    % Panel B: stratified by latent knowledge
    \begin{subfigure}{0.48\textwidth}
        \centering
        \includegraphics[width=\linewidth]{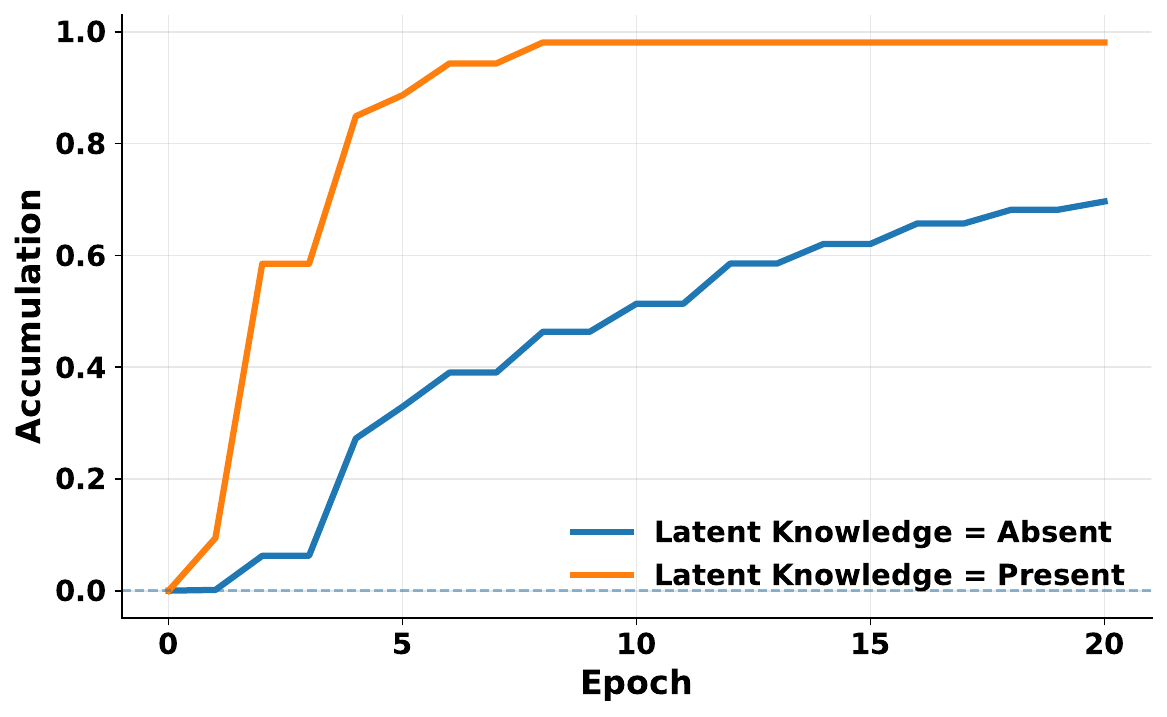}
        \caption{Accumulation stratified by latent knowledge.}
        \label{fig:KM-latent}
    \end{subfigure}

    \caption{Fact accumulation curves $F(t)$ for 800 HPO term–identifier pairs over 20 epochs.
    (A) Kaplan–Meier–style cumulative accumulation curve $F(t)$ inverted from the survival
    function $S(t)$; accumulation converges at 71.9\% at epoch 20 (shaded areas show 95\% CIs).
    (B) $F(t)$ stratified by latent knowledge status. At epoch 20, terms with latent knowledge
    are learned with 98.4\% accuracy versus 69.7\% for terms without latent knowledge.}
    \label{fig:hpo-accumulation-combined}
\end{figure}

\begin{figure}[htbp]
    \centering
    % Panel A: velocity, latent vs absent
    \begin{subfigure}{0.48\textwidth}
        \centering
        \includegraphics[width=\linewidth]{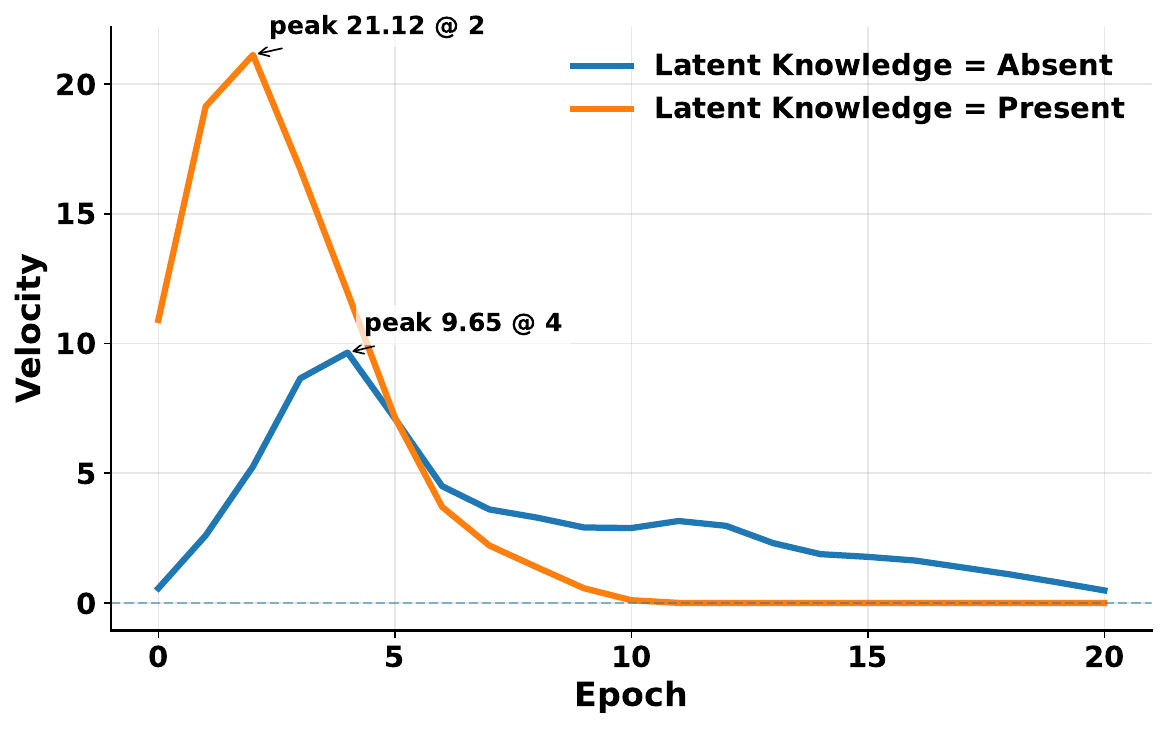}
        \caption{Accumulation velocity  for terms with latent knowledge present vs.\ absent. Peak velocity was 21.1\% per epoch at epoch~2 when latent knowledge was present and 9.7\% per epoch at epoch~4 when latent knowledge was absent.}
        \label{fig:velocity-hpo}
    \end{subfigure}
    \hfill
    % Panel B: accumulation, high vs moderate latent knowledge
    \begin{subfigure}{0.48\textwidth}
        \centering
        \includegraphics[width=\linewidth]{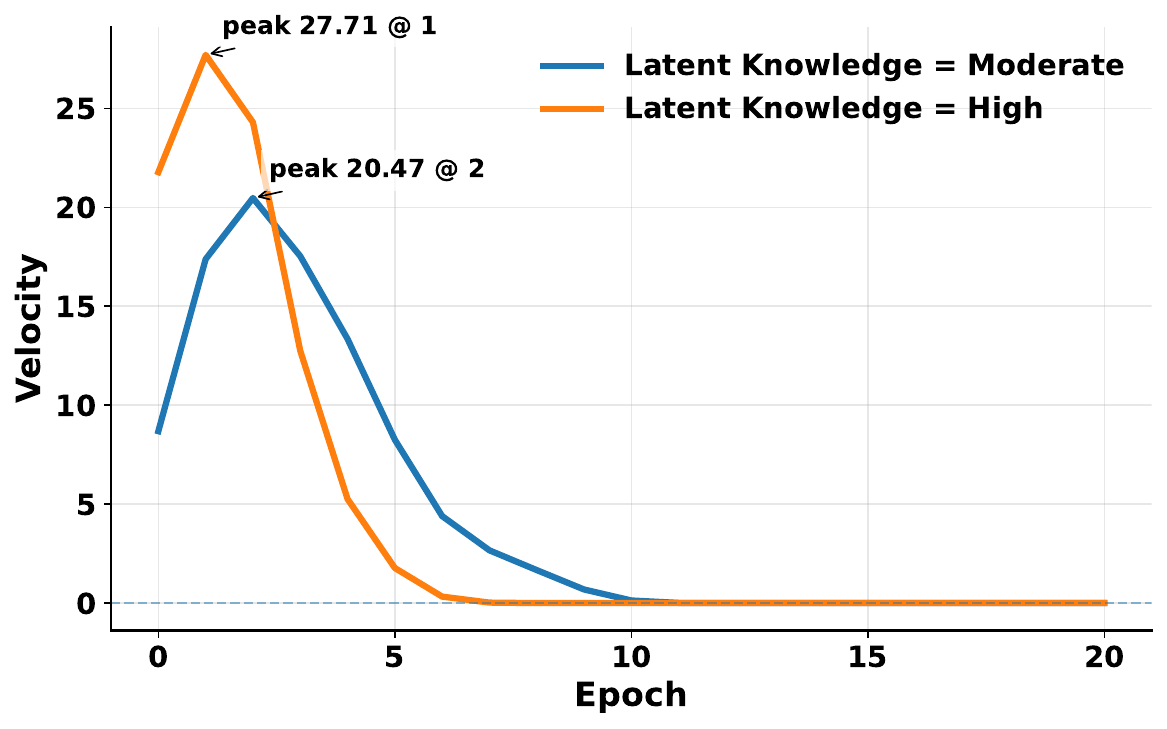}
        \caption{Fact accumulation curves for high latent knowledge terms (stochastic accuracy $\geq 0.10$) vs.\ moderate latent knowledge terms ($0 < $ stochastic accuracy $< 0.10$). More latent knowledge yielded higher velocities.}
        \label{fig:acquisition-high-latent-knowledge-terms}
    \end{subfigure}
    \caption{Dose–response relationship between latent knowledge and fact acquisition. 
    (A) Terms with any latent knowledge show earlier and higher peak accumulation velocities than terms without latent knowledge. 
    (B) Among terms with latent knowledge, higher stochastic accuracy is associated with earlier and more complete fact acquisition, consistent with a graded (dose–effect) influence of latent knowledge on learning dynamics.}
    \label{fig:hpo-latent-velocity-combined}
\end{figure}

\section{Results}
\subsection{Fact Accumulation for HPO Term–Identifier Pairs}
Fact accumulation for HPO term–identifier pairs across training epochs was summarized using the accumulation function \(F(t)\)
(Fig.~\ref{fig:km-overall-hpo}). \(F(t)\) increased steadily over the
20-epoch training period, with 71.9\% of term–identifier pairs correctly
acquired by epoch~20.  All 800 HPO terms were included in the training set
(i.e., all facts were \textit{seen}; none were withheld). Before any
fine-tuning (epoch~0), the model mapped 22 terms (2.8\%) correctly.  Among the remaining facts, the mean time to fact acquisition was \(10.8 \pm 0.3\) epochs.

\subsection{Effect of Latent Knowledge on Accumulation of Facts about HPO Terms During Fine-Tuning}
Accumulation curves for HPO term facts differed markedly by latent-knowledge status (Fig.~\ref{fig:KM-latent}). Terms for which the base model had latent knowledge were acquired earlier and faster than terms that lacked latent knowledge. Mean time to acquisition was \(2.9 \pm 0.4\) epochs for HPO terms with
latent knowledge versus \(11.4 \pm 0.3\) epochs for HPO terms without latent knowledge. The curves differed significantly (log-rank test: \(\chi^{2} = 212,\; df = 1,\; p < 0.001\)).

\subsection{Learning Velocity for Facts about HPO Terms is Predicted by Latent Knowledge}
The first derivative of the accumulation curve, \(V(t)\), is the rate
of fact acquisition per epoch (Fig.~\ref{fig:velocity-hpo}). Terms with latent knowledge reached a higher and earlier peak velocity (21.1\% per epoch at epoch~2) than terms without latent knowledge (9.7\% per epoch at epoch~4).  By epoch~10, the velocity for terms with latent knowledge had declined to zero,
indicating convergence. In contrast, velocity for terms without latent knowledge remained positive through epoch~20,  reflecting a slower and more prolonged learning trajectory and delayed convergence.

\subsection{Evidence of a dose-effect for Latent Knowledge for Fact Acquisition about HPO Terms}
Within the latent-knowledge subset, high–latent-knowledge terms were acquired
faster than moderate–latent-knowledge terms (mean \(1.0 \pm 0.4\) vs.
\(3.4 \pm 0.4\) epochs; Fig.~\ref{fig:acquisition-high-latent-knowledge-terms});
this difference was also significant (log-rank:
\(\chi^{2} = 18,\; df = 1,\; p < 0.001\)). 

\subsection{Predictors of Fact Acquisition About HPO Terms}
\begin{figure}[H]
    \centering
    % Panel A: HPO acquisition
    \begin{subfigure}{0.48\linewidth}
        \centering
        \includegraphics[width=\linewidth]{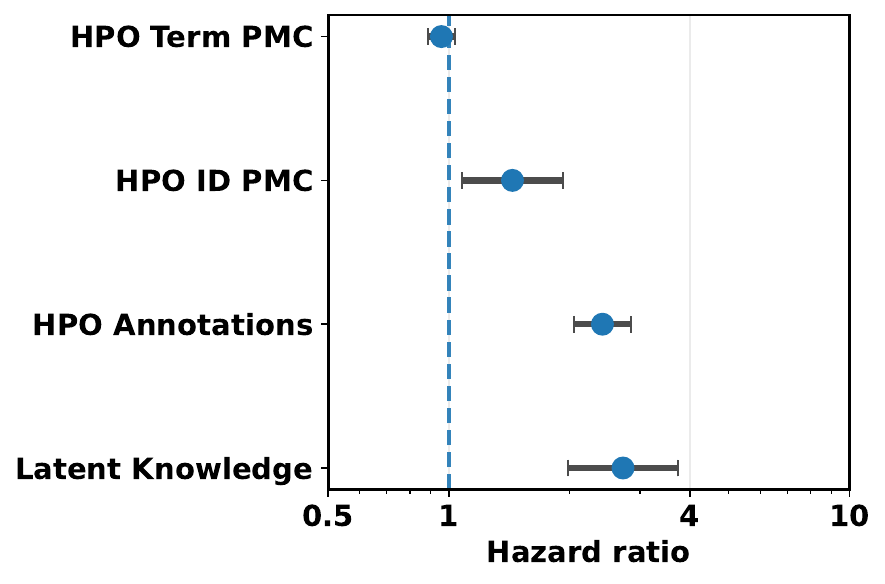}
        \caption{Covariates predicting fact acquisition for HPO term–identifier pairs.}
        \label{fig:forest_accumulation_hpo}
    \end{subfigure}
    \hfill
    % Panel B: GO generalization
    \begin{subfigure}{0.48\linewidth}
        \centering
        \includegraphics[width=\linewidth]{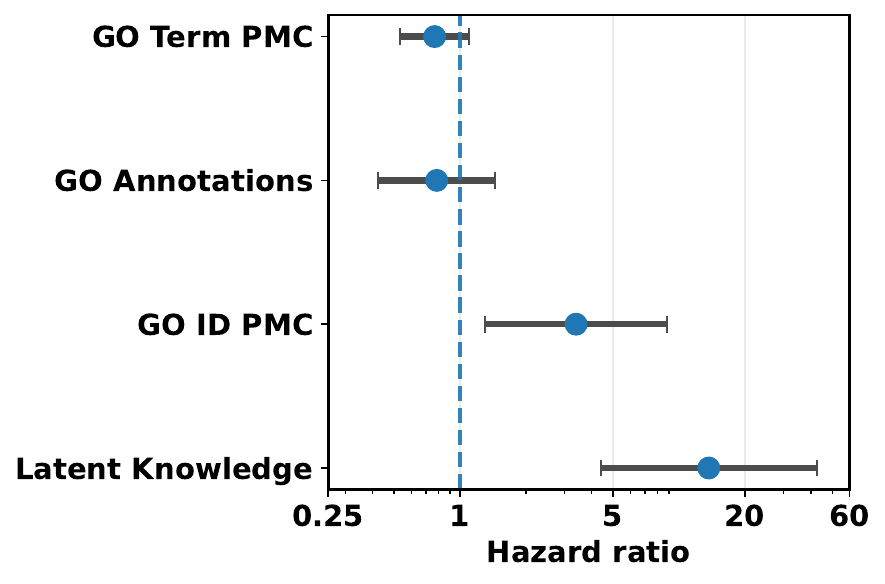}
        \caption{Covariates predicting generalization to \textit{unseen} GO term–identifier pairs.}
        \label{fig:forest_generalization_go}
    \end{subfigure}

    \caption{Cox proportional hazards models for (A) fact acquisition in HPO and (B) generalization to unseen GO term–identifier pairs during fine-tuning. Latent knowledge emerged as the strongest predictor of both fact acquisition for HPO terms and generalization for GO terms during fine-tuning.}
    \label{fig:forest_combined}
\end{figure}

Cox proportional hazards modeling (Fig.~\ref{fig:forest_combined}, Table~\ref{tab:cox-covariates}) identified latent knowledge as the strongest predictor of fact acquisition about HPO terms. The presence of latent knowledge increased the hazard of acquisition by approximately
160\% (hazard ratio \(= 2.6\), \(p < 0.001\)). Two popularity measures—annotation frequency and identifier frequency in PMC—were also significant positive predictors of  fact acquisition rate
\cite{kandpal2023large,ghosal2024understanding,christoph2025data}. In contrast, term frequency itself in the PMC was not significantly associated with the acquisition rate. Violations of the proportional hazards assumption (Schoenfeld residual
tests) indicated that covariate effects were not constant across epochs. Accordingly, the estimated hazard ratios should be interpreted as \textit{time-averaged effects} over the fine-tuning period rather than as strictly proportional over time.

\begin{table}[H]
\caption{Predictors of \textbf{Fact Acquisition} (Cox Proportional Hazards Model for Seen HPO Term–Identifier Pairs)}
\label{tab:cox-covariates}
\centering
\begin{tabular}{lrrr}
\toprule
Covariate & HR & 95\% CI & p-value \\
\midrule
{HPO term PMC}      & 0.96 & 0.89--1.04 & 0.30 \\
\textbf{HPO ID PMC}        & 1.44 & 1.08--1.93 & 0.01 \\
\textbf{HPO annotations}   & 2.42 & 2.06--2.85 & $<\!0.001$ \\
\textbf{Latent knowledge}  & 2.72 & 1.98--3.73 & $<\!0.001$ \\
\bottomrule
\end{tabular}
\begin{tablenotes}
\footnotesize
\item \textbf{Note.} HR = hazard ratio. HR $>$ 1 indicates a higher rate of fact accumulation; HR $<$ 1 indicates a lower rate. Latent knowledge has the highest HR and is the strongest predictor of learning velocity. Bolded covariates are statistically significant $p < 0.01.$
\end{tablenotes}
\end{table}

\begin{table}[H]
\caption{Predictors of \textbf{Generalization} (Cox Proportional Hazards Model for Unseen GO Term–Identifier Pairs)}
\label{tab:cox-generalization}
\centering
\begin{tabular}{lrrr}
\toprule
Covariate & HR & 95\% CI & p-value \\
\midrule
{GO term PMC}        & 0.77 & 0.53--1.10 & 0.15 \\
\textbf{GO ID PMC}          & 3.39 & 1.30--8.83 & 0.01 \\
{GO annotations}     & 0.78 & 0.42--1.45 & 0.44 \\
\textbf{Latent knowledge}   & 13.67 & 4.39--42.59 & $<\!0.005$ \\
\bottomrule
\end{tabular}
\begin{tablenotes}
\footnotesize
\item \textbf{Note.} HR = hazard ratio. HR $>$ 1 indicates a higher rate of fact generalization; HR $<$ 1 indicates a lower rate. Latent knowledge shows the largest HR and is the strongest predictor of generalization risk among unseen GO terms. Bolded covariates are statistically significant $p < 0.01.$
\end{tablenotes}
\end{table}
\begin{table}[H]
\caption{Predictors of \textbf{Degradation} (Cox Proportional Hazards Model for Initially Correct GO Term–Identifier Pairs)}
\label{tab:cox-degradation}
\centering
\begin{tabular}{lrrr}
\toprule
Covariate & HR & 95\% CI & p-value \\
\midrule
{GO term PMC}        & 1.04 & 0.60--1.80 & 0.88 \\
{GO ID PMC}          & 0.33 & 0.10--1.16 & 0.08 \\
{GO annotations}     & 0.62 & 0.27--1.42 & 0.26 \\
{Latent knowledge}   & 0.86 & 0.16--4.65 & 0.86 \\
\textbf{Seen during training}   & 0.10 & 0.03--0.32 & $<\!0.005$ \\
\bottomrule
\end{tabular}
\begin{tablenotes}
\footnotesize
\item \textbf{Note.} HR = hazard ratio. HR $>$ 1 indicates a higher rate of degradation; HR $<$ 1 indicates a lower rate. Being a seen (trained) GO term was associated with a substantially lower hazard of degradation compared with unseen terms. Bolded covariates are statistically significant $p < 0.01.$
\end{tablenotes}
\end{table}

\subsection{Latent Knowledge Predicts Generalization to Unseen GO Terms during Fine-Tuning}
We next asked whether fine-tuning could induce \textit{generalization} to GO term–identifier pairs that were never seen during training. For this analysis,  we examined GO terms that were held out from the fine-tuning set and that were incorrect at baseline (epoch 0). This yielded 379 \textit{unseen} GO terms for evaluation. Generalization was defined as the first epoch in which the model produced the correct GO identifier for an \textit{unseen}~term. Terms that never became correct by epoch 20 were right-censored. 

Generalization to \textit{unseen} GO terms was uncommon but not entirely absent. Over 20 epochs, 22 of 379 terms (5.8\%) became correct, with the Kaplan--Meier curve showing that the probability of remaining incorrect decreased from 1.0 at baseline to approximately 0.94 at epoch 20. Among the small subset of generalizing terms, the median time-to-generalization was 3 epochs (interquartile range 2–7), indicating that successful generalization occurred early in training rather than gradually accumulating over later epochs.

Although uncommon, generalization was predictable from properties of the model and the identifiers. A Cox proportional hazards model fitted to the \textit{unseen} GO terms (concordance = 0.90, log-likelihood ratio test $p < 10^{-10}$) identified latent knowledge and identifier popularity as significant predictors of time-to-generalization (Table \ref{tab:cox-generalization}). Latent knowledge was operationalized as whether the model could ever produce the correct identifier for a term under high-temperature stochastic decoding prior to fine-tuning. \textit{Unseen} terms with evidence of latent knowledge were much more likely to generalize (hazard ratio = $13.67$) than terms with no such evidence. In addition, higher GO identifier frequency in PMC was associated with increased generalization (hazard ratio = $3.39$), consistent with the idea that facts supported by more textual evidence in pretraining are easier to consolidate. In contrast, GO annotation counts and GO term frequencies in PMC did not reach significance after controlling for these effects. 

Taken together, these results suggest that fine-tuning can occasionally extend knowledge to \textit{unseen} GO term–identifier pairs, but such generalization is uncommon and systematically constrained. It occurs primarily for terms that already have latent parametric support and for identifiers that are well represented in biomedical text, underscoring the central role of pretraining-derived structure in governing where factual generalization is possible.

\begin{figure}[H]
    \centering
    % Panel (a): overall generalization
    \begin{subfigure}{0.48\linewidth}
        \centering
        \includegraphics[width=\linewidth]{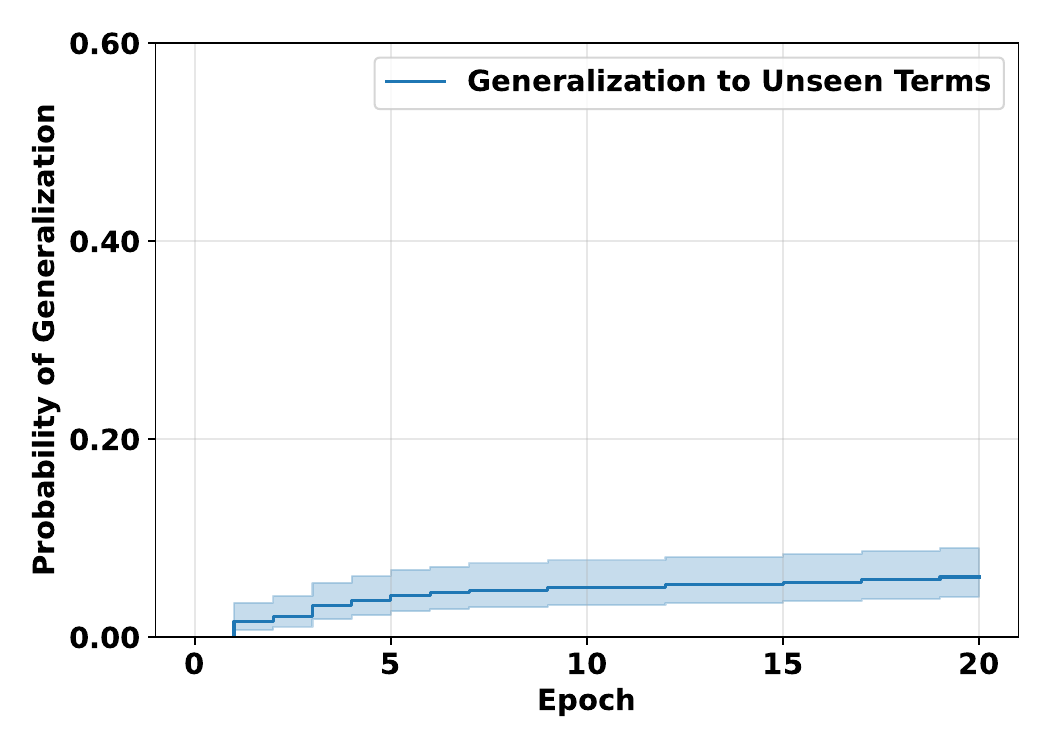}
        \caption{Fact learning generalized to 5.8\% of \textit{unseen} GO terms evaluated after fine-tuning. Models were trained on the \textit{seen} terms and tested on the 379 \textit{unseen} terms. 21 GO terms that were correctly mapped at baseline were excluded.}
        \label{fig:generalization-all}
    \end{subfigure}
    \hfill
    % Panel (b): generalization by latent knowledge
    \begin{subfigure}{0.48\linewidth}
        \centering
        \includegraphics[width=\linewidth]{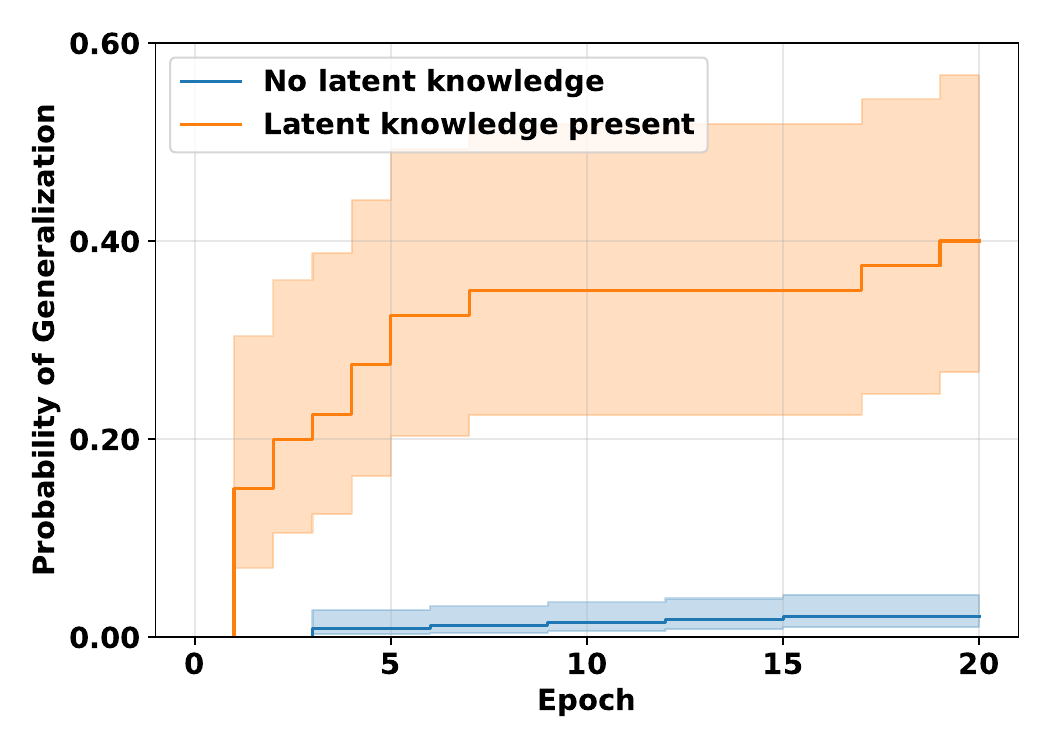}
        \caption{Generalization in \textit{unseen} terms without latent knowledge (blue line, $n=338$) was negligible. Latent knowledge in \textit{unseen} terms was uncommon ($n=40$) but associated with a substantially higher probability of generalization (orange line).}
        \label{fig:generalization-by-latent}
    \end{subfigure}
    
    \caption{Generalization to \textit{unseen} GO term–identifier pairs during fine-tuning. (a) Overall cumulative probability of generalization across all \textit{unseen} terms. (b) Stratification by latent knowledge shows that generalization is concentrated in the subset of \textit{unseen} terms supported by latent knowledge.}
    \label{fig:generalization-go}
\end{figure}

\subsection{Training Protects Against Degradation of Previously Correct GO Term Mappings During Fine-Tuning}

To evaluate whether fine-tuning disrupts previously correct knowledge, we analyzed the 43 GO term–identifier pairs that were already correct at baseline (epoch 0). During fine-tuning, 24 of these 43 terms (55.8\%) degraded—that is, the model began producing an incorrect identifier for a fact it previously knew.

\begin{figure}[H]
    \centering
    % Panel (a): all terms
    \begin{subfigure}{0.48\linewidth}
        \centering
        \includegraphics[width=\linewidth]{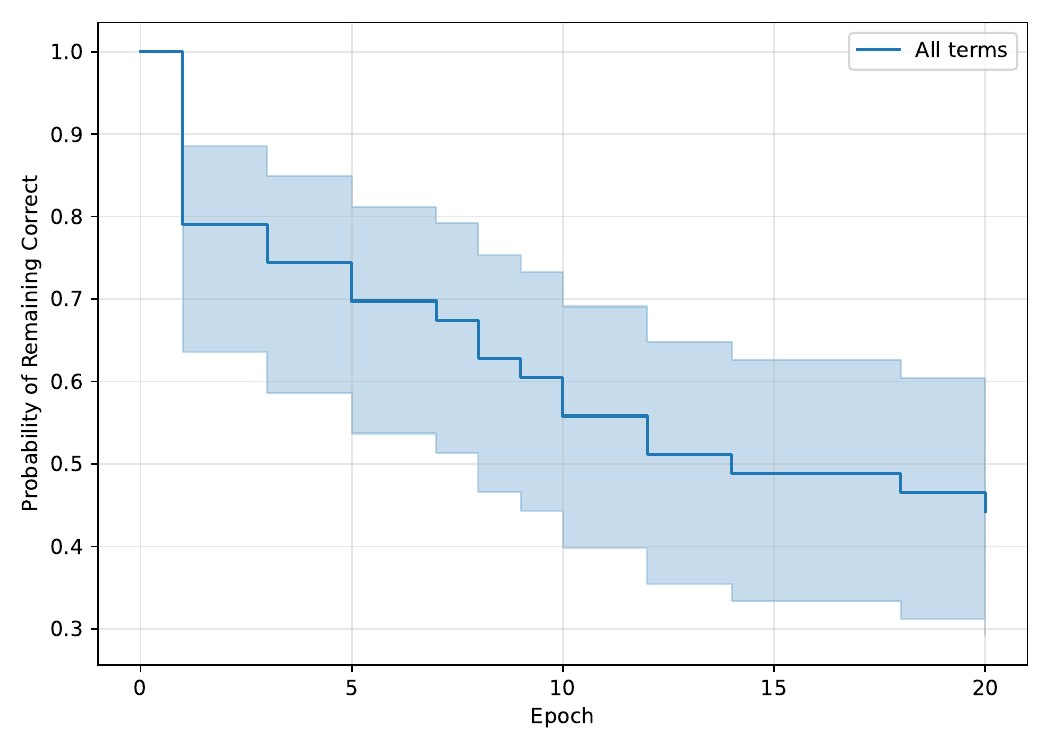}
        \caption{Fine-tuning on the \textit{seen} terms had adverse effects on terms already known by the model prior to fine-tuning, consistent with prior reports of knowledge loss during fine-tuning. Shaded regions indicate 95\% confidence intervals for the Kaplan--Meier survival curves.}
        \label{fig:degradation-all}
    \end{subfigure}
    \hfill
    % Panel (b): seen vs unseen
    \begin{subfigure}{0.48\linewidth}
        \centering
        \includegraphics[width=\linewidth]{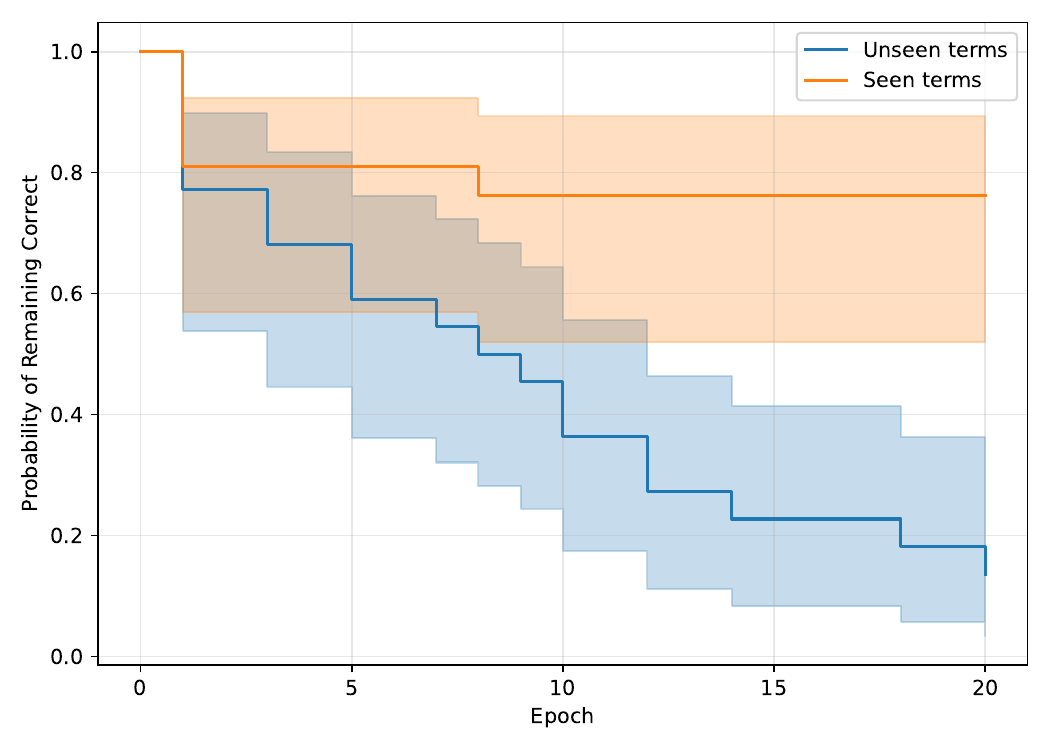}
        \caption{Degradation was more rapid for \textit{unseen} GO terms (blue line) than for \textit{seen} GO terms (orange line), suggesting a protective effect of training against degradation of pre-existing knowledge. Shaded regions indicate 95\% confidence intervals for Kaplan-Meier plots.}
        \label{fig:degradation-seen-unseen}
    \end{subfigure}
    
    \caption{Degradation of GO term–identifier mappings during fine-tuning. (a) Overall degradation across all terms known at baseline. (b) Stratification by \textit{seen} versus\textit{ unseen} GO terms shows a protective effect of training on degradation.}
    \label{fig:degradation-go}
\end{figure}

Figure \ref{fig:degradation-go} the Kaplan–Meier survival curves stratified by training status. Degradation occurred substantially more often in \textit{unseen} terms than in \textit{seen} terms that had been included in the fine-tuning set. The Mantel–Cox log-rank test confirmed a robust difference in survival distributions $(\chi^2 = 14.47,\ p = 1.4 \times 10^{-4})$, indicating that fine-tuning provides protective reinforcement for the facts on which it is trained.

We used a Cox proportional hazards model to identify predictors of degradation. Among the covariates evaluated—latent knowledge score, GO annotation frequency, GO identifier frequency, and GO term frequency—only the training status of \textit{unseen} significantly predicted degradation rate (Table \ref{tab:cox-degradation}). Terms included in fine-tuning were 90\% less likely to degrade than \textit{unseen} terms (hazard ratio = $0.10$). None of the pretraining-derived measures (latent knowledge or popularity in the corpus) showed a significant predictive effect for degradation during fine-tuning on the GO terms.

These findings indicate that degradation during fine-tuning is common but predictable: facts excluded from the fine-tuning dataset are substantially more vulnerable to catastrophic forgetting, whereas trained facts are protected.

\section{Discussion}

This study provides a framework for viewing fact acquisition during fine-tuning
of large language models as a rate-based process rather than a binary outcome.
Using Human Phenotype Ontology (HPO) term–identifier pairs as a structured set
of learnable facts~\cite{Hier2025failures, pericharla2025memorizationgeneralizationfinetuninglarge},
we characterized how rapidly new facts are acquired during fine-tuning and how
these rates depend both on the properties of individual facts and on the
latent knowledge present in the pretrained model.

This framework helps explain why binary accuracy metrics can obscure changes in
model performance after fine-tuning. Two fine-tuned models with similar final
performance (accuracy) may reach that state through different mixtures of early
memorization, delayed acquisition, and loss of previously correct facts.
Pretraining-induced representations of facts in the model (often described as
factual salience) play a central role in this process: fact acquisition during
fine-tuning is strongly influenced by what was partially encoded in the model
parameters prior to supervision~\cite{ghosal2024understanding}.

Latent knowledge—information present in the pretrained model but not reliably
accessible through deterministic decoding—was the strongest predictor of rapid
fact acquisition \cite{gekhman2024does, gekhman2025inside}. HPO terms with latent knowledge were learned more accurately (100\% vs.\ 60\%), showed higher early learning velocities (21.1\% vs.\ 9.7\%
per epoch), and reached their peak learning rates earlier (epoch~2 vs.\ epoch~4).
They also converged sooner: learning velocity declined to zero by epoch~10 for
terms with latent knowledge, whereas terms without latent knowledge continued
to be acquired through epoch~20. Within the latent-knowledge subset,
high–latent-knowledge terms converged earlier (around epochs~4–8), reached
higher peak velocities (27.8\% vs.\ 20.5\% per epoch), and achieved those peaks
earlier (epoch~1 vs.\ 2) than moderate–latent-knowledge terms
(Fig.~\ref{fig:hpo-latent-velocity-combined}).

It is important to emphasize that latent knowledge is not a property of the
ontology term itself but of the \textit{model–term pair}. Each pretrained model
(e.g., Llama-3.1-8B-Instruct, Llama-3.1-70B, Llama-3.1-405B) will exhibit its own
repertoire of latent knowledge about ontology terms, determined by its
pretraining corpus and internal parameterization. A term can display latent
knowledge in one model and be unknown to another. Thus, latent knowledge is
 model--specific and term--specific and is not an intrinsic attribute of the
term. This asymmetry highlights a structural limitation of fine-tuning as a
knowledge-injection strategy: remediation is most effective when it amplifies
weak but present signals, and less effective when representational capacity
must be allocated entirely anew, as in \textit{ontology desert} terms that are
essentially absent from pretraining corpora~\cite{Hier2025failures}.

Although we did not examine the model’s search space directly, our findings
suggest that latent knowledge likely narrows the model’s effective search space to a
smaller set of high-probability pairings during fine-tuning. When a term
already has a weak internal association with its identifier, fine-tuning may
\textit{adjust} existing next-token probabilities rather than discovering the
association from scratch.  We suggest that latent knowledge gives the model a
\textit{head start} in its search for the identifier that correctly matches the
ontology term. A narrowed search space could help explain two characteristic
signatures observed in our results: (1) higher initial learning velocities and
(2) faster convergence, with velocity reaching zero by approximately epoch~10
for HPO terms with latent knowledge. In contrast, terms lacking latent
knowledge start with more diffuse next-token probabilities across many possible
identifiers and must first \textit{discover} a plausible association through
gradual gradient updates \cite{speicher2024memorisation}. This is consistent with (1) lower initial velocities
and (2) prolonged accumulation, with velocity approaching zero only by
epoch~20 for such terms.

Under this interpretation, which remains somewhat speculative, fact accumulation is driven primarily by a redistribution of next-token probability mass, with fine-tuning shifting probability from many low-plausibility identifiers toward the correct one~\cite{chang2024large}. Other mechanisms—such as embedding realignment or changes in decision boundaries—may also contribute, but we do not examine them directly here~\cite{yavas2025relation, rajaee2021does, zhao2024probing}. Our analysis is restricted to observable learning curves rather than detailed mechanistic explanations.

The GO experiments extend these findings beyond HPO and allow us to examine
fact-based generalization and degradation during fine-tuning. For \textit{unseen} GO
terms withheld from the training set, fact-level generalization was uncommon but not
completely absent. Across 20 epochs of fine-tuning, 22 of 379 \textit{unseen} terms (5.8\%) changed from incorrect to correct, indicating that fine-tuning can induce correct new GO
mappings in terms that were excluded from training. These few terms that
did generalize (became correct during fine-tuning without direct exposure in training) were strongly enriched for latent knowledge. In the Cox model of time-to-generalization, the presence of latent knowledge about a term in the base model (high-temperature correctness at baseline) increased the hazard of generalization by more than an order of magnitude (hazard ratio= 13.67), and higher GO identifier frequency in PubMed Central provided an additional, smaller boost (hazard ratio = 3.39). \textit{Unseen} terms without latent knowledge showed essentially flat cumulative generalization curves, whereas high–latent-knowledge terms, though uncommon, exhibited a nontrivial probability of becoming correct during fine-tuning (Figure~\ref{fig:generalization-by-latent}). In a fact-acquisition paradigm, generalization therefore appears as a selective phenomenon concentrated in the small set of \textit{unseen} terms that the model already “half knows,” in the sense that weak but detectable latent associations
are present before fine-tuning. Lu et al.~\cite{lu2024scaling} also report that,
during pre-training on large collections of atomic facts, LLMs can generalize to
some \textit{unseen} fact triples, especially when there is a strong correlation
structure between inputs (query) and outputs (response), and that the resulting generalization
follows a scaling law similar to standard pre-training. Together, these findings
suggest that both pre-training and fine-tuning support fact-level generalization
only when the model’s representations already encode a non-trivial amount of
association between the fact input and the fact output, what we term as latent knowledge about the fact.

We also observed that fine-tuning carries a measurable cost in the form of
knowledge degradation~\cite{pletenev2025much}. Among 43 GO term–identifier pairs that were mapped correctly at baseline (epoch~0), more than half (24/43) became incorrect at
least once during the 20-epoch fine-tuning run. Degradation affected both \textit{seen}
and \textit{unseen} terms, but it was more rapid for \textit{unseen} GO terms than
for \textit{seen} terms. The Kaplan–Meier curves showed earlier and steeper declines in
the probability of remaining correct for \textit{unseen} terms, and a log-rank test
confirmed a significant difference between the groups ($\chi^2 = 14.47$,
$p = 1.4\times 10^{-4}$). In a multivariable Cox model that included latent
knowledge and corpus-based popularity measures, only absence from the training set
remained a strong predictor of degradation: \textit{seen} GO terms had an approximately
tenfold lower hazard of becoming incorrect than \textit{unseen} ones (hazard ratio = 0.10). Latent knowledge, which strongly facilitated fact acquisition,
did not  protect against fact degradation once a mapping was already
correct.

Taken together, these results highlight an asymmetric role for pretraining
structure during fine-tuning. Latent knowledge in the base model identifies
which \textit{unseen} ontology facts are most likely to benefit from fine-tuning and to
generalize, whereas resistance to degradation depends primarily on whether a
fact is reinforced during training. This pattern is consistent with prior
reports that fine-tuning can improve performance in targeted domains while
eroding unrelated or weakly supported knowledge, implying a trade-off between
knowledge injection and knowledge preservation~\cite{pletenev2025much}. In
practical terms, our findings support fine-tuning regimes that (1) monitor
fact-level degradation, (2) prioritize reinforcement of clinically important
facts that are already correct, and (3) use latent-knowledge probes to identify
those \textit{unseen} ontology terms where fine-tuning is most likely to yield
true generalization rather than noise.

\subsection{Implications for Biomedical Informatics}
Although our experiments focus on HPO and GO term–identifier mappings, the framework applies to any ontology-grounded fact with a verifiable term–code association, including SNOMED CT concepts, RxNorm identifiers, LOINC codes, and ICD codes.  Rate-based characterizations of fact acquisition provide information that binary accuracy cannot as they quantify which facts are learned rapidly, which require extended training, and which remain  unlearned within a fixed fine-tuning budget.

These measurements have several direct informatics applications. First, rate-based metrics enable systematic comparison of fine-tuning curricula, such as sampling strategies that emphasize long-tail or underrepresented terms, mixtures of domain-specific and general corpora, or alternative regularization schemes. Second, hazard ratios quantify how strongly factors such as latent knowledge or identifier prevalence influence acquisition speed, allowing developers to determine when supervised fine-tuning is likely to succeed and when alternative strategies—such as retrieval-augmented generation or targeted knowledge editing—may be preferable. Third, the framework offers a principled way to estimate how much training is required to inject specific ontology codes into a model, which can guide the construction of terminology-aware clinical LLMs.

Finally, because it reveals which terms exhibit minimal latent knowledge, the framework provides an explicit method for identifying \textit{ontology deserts}—terms from large biomedical ontologies that are structurally well defined but essentially absent from text corpora and therefore receive little or no representational capacity during pretraining~\cite{Hier2025failures}. Rate-based analysis can then be used to estimate how much fine-tuning would be required to populate these desert regions and to flag cases where alternative mechanisms are likely needed to ensure adequate coverage. In practical terms, our results distinguish terms that can be acquired efficiently by fine-tuning (those with latent knowledge in the model parameters) from terms in ontology deserts, for which fine-tuning is likely to be data-inefficient because the model lacks internal representations to support acquisition.

\subsection{Limitations}
This study has some limitations that should be considered when interpreting the findings. We operationalize factual acquisition using the task of learning the identifier associated with a biomedical ontology term. While alternative formulations of factual knowledge exist, the term–identifier paradigm offers important methodological advantages for this work: correctness can be assessed unambiguously, all facts share a common structural form, and large-scale annotation frequencies provide an empirically grounded proxy for fact popularity in the pretraining corpora. Nonetheless, the extent to which the observed learning rates generalize to other types of biomedical facts warrants further investigation.

The fine-tuning experiments were conducted using a fixed hyperparameter configuration while varying the number of training epochs (0–20) on datasets of 800 HPO terms or 400 GO terms. Other hyperparameter choices, including learning rate, batch size, and optimization strategy, were not explored. Prior work suggests that substantially longer training regimes—up to 100 epochs—may enable near-complete acquisition of ontology terms under some conditions~\cite{wang2023fine}. Accordingly, the acquisition rates reported here should be interpreted as conditional on the selected training configuration rather than as upper bounds on learnability.

We used a Cox proportional hazards model to identify covariates that predicted fact acquisition, generalization, and degradation during fine-tuning (Tables \ref{tab:cox-covariates} to \ref{tab:cox-degradation}).  Since HRs were not constant over time, the reported values represent average effects and may mask interactions with the training epoch.

Our analysis focused on changes in model accuracy and acquisition timing, rather than on the internal mechanisms that drive these changes. We did not directly examine shifts in logits, parameter updates, or representational geometry during fine-tuning. As a result, the study characterizes what is learned and when, but not the mechanistic processes by which internal representations are reorganized~\cite{ ghosal2024understanding, yavas2025relation,rajaee2021does,  zhao2024probing,feng2024extractive}. Integrating mechanistic analyses with the proposed time-to-event framework is a direction for future work.

Finally, our approach is tailored to fact acquisition tasks in which learning can be defined as a discrete event (e.g., the first epoch at which a correct identifier is produced). Extending this approach to skill-based domains such as reasoning or multi-step tool use is less straightforward, as these tasks lack a single, well-defined success criterion and would require adaptation to continuous or multi-dimensional learning signals. In addition, the method is computationally intensive: estimating latent knowledge requires repeated stochastic-decoding queries per term~\cite{gekhman2025inside}, and survival analysis necessitates evaluation of the full test set at each fine-tuning epoch.

\section{Conclusions}
We introduce a survival-analysis framework for assessing factual learning during model fine-tuning at the level of individual ontology facts. This approach quantifies both the \textit{rate} and the \textit{success} of fact acquisition, providing a richer description of fact learning than accuracy alone. Using ontology term–identifier pairs from HPO and GO as structured test cases, we model three outcomes during fine-tuning: memorization (acquisition of \textit{seen} facts), generalization (acquisition of \textit{unseen} facts), and degradation (loss of previously correct facts).

Our HPO experiments show that facts supported by latent knowledge acquired during pretraining are learned more rapidly and more completely, with clear dose–response effects on acquisition velocity. In GO, latent knowledge and identifier popularity predict which \textit{unseen} term–identifier pairs are most likely to generalize during fine-tuning, although such generalization events remain uncommon. By contrast, we find no evidence that latent knowledge protects against degradation of previously correct facts; the main protective factor is continued exposure during training.

These findings suggest that pretraining-derived knowledge narrows the model’s effective search space, increasing the efficiency with which factual associations are acquired during fine-tuning. The proposed rate-based framework can help guide future work on fine-tuning strategies, curriculum design, and evaluations of learning efficiency in large language models, particularly for ontology-grounded biomedical applications.

\section*{Acknowledgements}
None.

\section*{Conflicts of Interest}
The authors declare no conflicts of interest.

\section*{Human Studies}
In this study protected health information was not used and IRB approval was not required.

\section*{Funding}
The authors received no external funding for this work.

\section*{CRediT Author Statement}
Conceptualization: Daniel Hier, Tayo Obafemi-Ajayi; Methodology: Daniel Hier, Tayo Obafemi-Ajayi; Writing—original draft: Daniel Hier; Writing—review \& editing: Daniel Hier, Tayo Obafemi-Ajayi.

\section*{Declaration of Generative AI and AI-Assisted Technologies}

During the preparation of this manuscript, the authors used generative AI tools (GPT-5.1, OpenAI) to assist with phrasing and language refinement. After using these tools, the authors carefully reviewed and edited all content and accept full responsibility for the final manuscript.

\section*{Data and Code Availability}

Data and python code are available on Zenodo \cite{hier_2025_velocity_data}

\bibliographystyle{elsarticle-num}
\bibliography{references}

\end{document}